\documentclass{article}

\usepackage[preprint, nonatbib]{nips_2018}

\usepackage[utf8]{inputenc} 
\usepackage[T1]{fontenc}    
\usepackage{hyperref}       
\usepackage{url}            
\usepackage{booktabs}       
\usepackage{amsfonts}       
\usepackage{nicefrac}       
\usepackage{microtype}      
\usepackage{amsmath}
\usepackage{amssymb}

\usepackage{graphicx}
\usepackage{siunitx}

\usepackage[font=footnotesize,labelfont=footnotesize]{subcaption}
\usepackage[font=footnotesize,labelfont=footnotesize]{caption}

\DeclareMathOperator*{\argmax}{arg\,max}

\title{Active Learning with Gaussian Processes for High Throughput Phenotyping}

\author{
  Sumit Kumar\\
  Carnegie Mellon University\\
  \texttt{sumitsk@cmu.edu} \\
  \And
  Wenhao Luo\\
  Carnegie Mellon University\\
  \texttt{wenhaol@cs.cmu.edu} \\
  \And 
  George Kantor\\
  Carnegie Mellon University\\
  \texttt{kantor@ri.cmu.edu} \\
  \And 
  Katia Sycara\\
  Carnegie Mellon University\\
  \texttt{katia@cs.cmu.edu} \\
}

\begin{document}

\maketitle

\begin{abstract}
A looming question that must be solved before robotic plant phenotyping capabilities can have significant impact to crop improvement programs is scalability. High Throughput Phenotyping (HTP) uses robotic technologies to analyze crops in order to determine species with favorable traits, however, the current practices rely on exhaustive coverage and data collection from the entire crop field being monitored under the breeding experiment. This works well in relatively small agricultural fields but can not be scaled to the larger ones, thus limiting the progress of genetics research. In this work, we propose an active learning algorithm to enable an autonomous system to collect the most informative samples in order to accurately learn the distribution of phenotypes in the field with the help of a Gaussian Process model. We demonstrate the superior performance of our proposed algorithm compared to the current practices on sorghum phenotype data collection. 
\end{abstract}

\section{Introduction}
World population is projected to reach $9.6$ billion by $2050$, and yields of most staple crops are not increasing at a fast enough rate to meet the corresponding nutritional needs. To make matters worse, the total cultivable landmass is more likely to decline due to a number of factors like ever-increasing urbanization, non-biodegradable waste generation, etc. Hence, it is important to find genetic varieties of crops with favorable traits like disease resistance, high yield, etc. via plant breeding processes~\cite{furbank2011phenomics} to ensure food security for the next generations. 
Plant phenotyping evaluates crops based on physical characteristics to support plant breeding and genetic activities. Since the current standard practice in collecting phenotype data needs human specialists to assess thousands of plants, phenotyping is currently the bottleneck in the breeding process. To increase efficiency, High Throughput Phenotyping (HTP) uses sensors and robotic platforms to gather phenotype data~\cite{mueller2017robotanist}. 

Current field phenotyping systems rely on exhaustive coverage of the crop field being studied in the breeding experiment. This works well in relatively small proof-of-concept trials but requires significant time and resources for a large field which in turn limits the progress of genetics research. This is especially true for ground-based platforms which can collect higher quality data than aerial platforms, but have much lower coverage rates. 
With the aim to accelerate the breeding process, we propose an active learning algorithm for an autonomous ground robot to collect a subset of samples with high utility in a short time. The agent models the distribution of a target phenotype of interest in the field with a Gaussian Process and actively selects the next locations to sample instead of exhaustively visiting the whole field.

In HTP, phenotypes like plant height, stalk width, etc. are measured with the help of state-of-the-art computer vision and deep learning techniques~\cite{jenkins2017online, baweja2018stalknet, kayacan2018embedded} on images captured by on-board cameras. Note that images can be captured by the robot even when it is in motion. This means that as the robot is travelling from one place to another, it can gather phenotype measurements from all the sampling locations or plots along the path. This is different from conventional adaptive sampling where the robot selects some locations to collect high utility data and plans a path to reach them by optimizing a cost function like path length. In this model, the robot learns about the distribution of a target variable of interest by taking measurements at the selected locations and there is no information gain as the robot is travelling from one location to another as it is not taking any measurements along the way.

\begin{figure*}
    \centering
        \begin{minipage}{0.33\textwidth}
        \centering
        \includegraphics[width=0.95\textwidth]{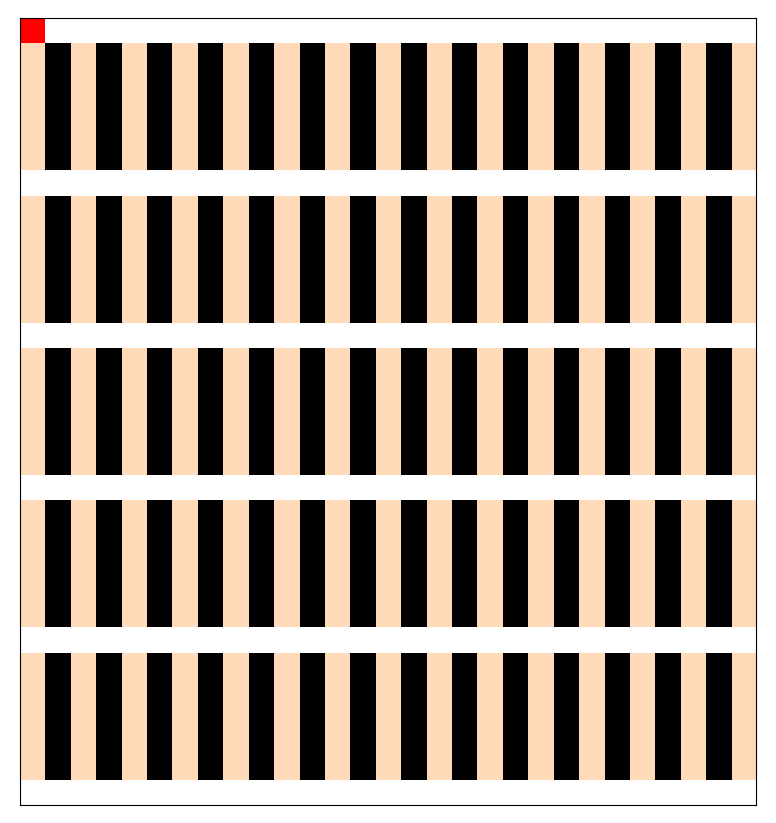} 
        \subcaption{Environment}
        \label{fig:grid_env}
    \end{minipage}\hfill
    \begin{minipage}{0.33\textwidth}
        \centering
        \includegraphics[width=0.95\textwidth]{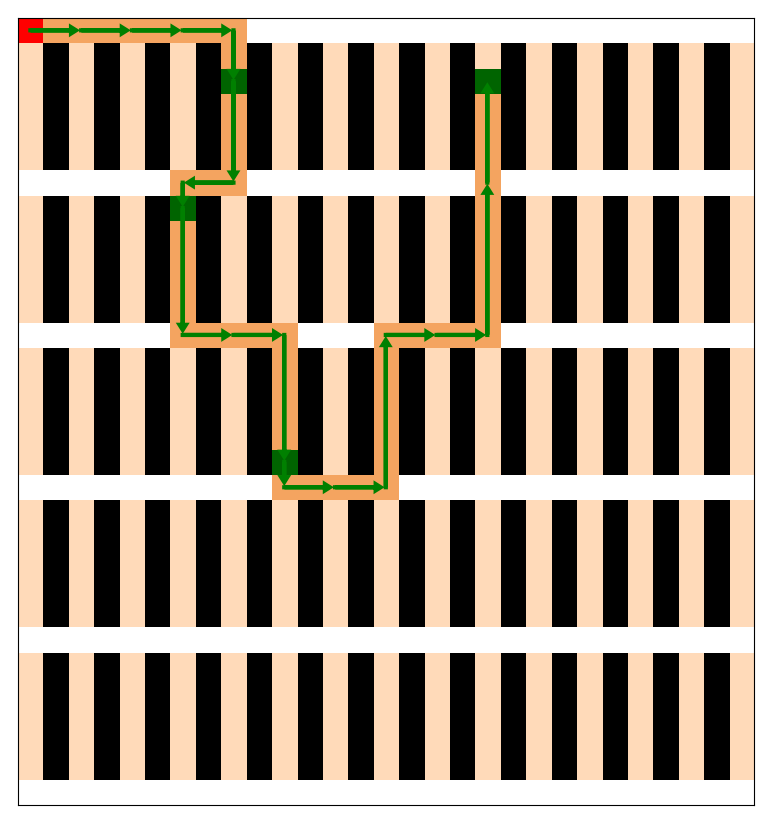} 
        \subcaption{Iteration 1}
        \label{fig:iter1}
    \end{minipage}
    \begin{minipage}{0.33\textwidth}
        \centering
        \includegraphics[width=0.95\textwidth]{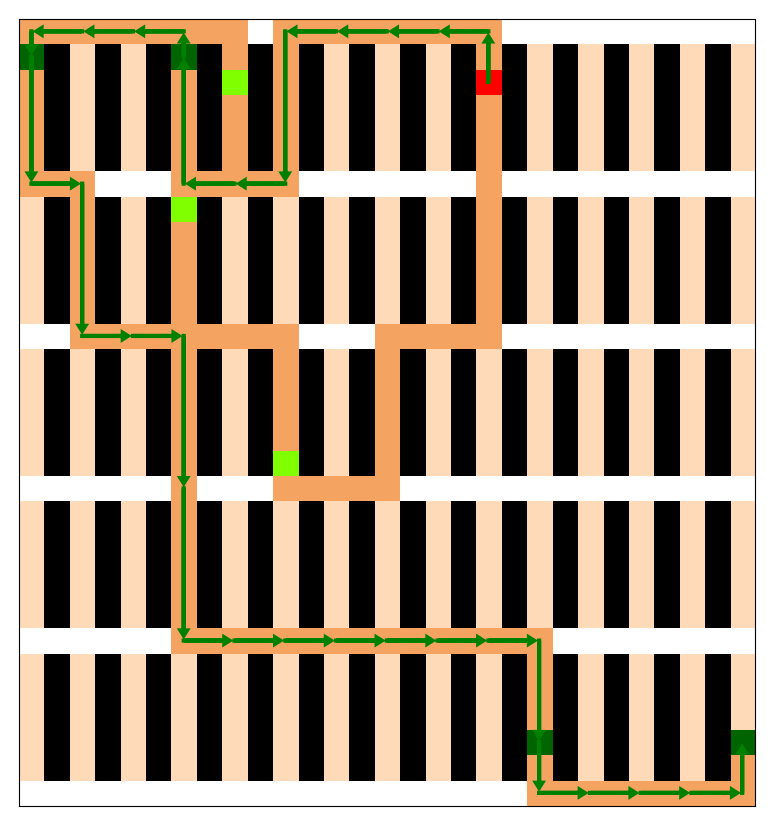} 
        \subcaption{Iteration 2}
        \label{fig:iter2}
    \end{minipage}
    \caption{(a) The grid environment used in simulation. The red cell depicts the current position of the robot. The black cells are obstacles whereas the rest is free space. The sampling locations or plots in the field are shown in orange. (b) The robots determines some most informative sensing locations (shown as dark green cells) to collect high quality phenotype measurements. It then plans a path from its current position to visit all the sampling locations. As the robot traverses the planned path (shown as a sequence of green arrows), it gathers noisy measurements from all the plots along the way. (c) After completing the traversal, it repeats this process from its current position. For picture clarity, all the static samples collected in previous iterations are shown in light green color. The dark orange colored regions are the areas visited by the agent so far.}\label{fig:env}
\end{figure*}

In contrast, in this work, we consider the case where the robot collects data (through images) not only at some selected locations but also while moving along the planned path in the field as shown in Figure \ref{fig:env}. Formally, there are two types of measurements: 
\begin{itemize}
    \item \textit{Static} measurements: The robot stops at a sampling location or plot in the field and takes images after orienting itself properly with respect to the plants in order to minimize the measurement noise and gather highly accurate data. 
    \item \textit{Mobile} measurements: As the robot is travelling from one point to another in the field, it takes phenotype measurements from the images captured from all the plots along the path without any adjustments.
\end{itemize}

Naturally, the static measurements are more accurate than the mobile ones due to localization errors and noisy images resulting from robot's motion on the uneven terrains of the field. In fact, the larger the speed of the robot, the noisier is the mobile data. However, gathering static measurements requires more time and resources than mobile ones. As a result, there is a trade-off between the quality of data and the required resources. Hence, it is important to develop statistical machine learning models and adaptive sampling algorithms to incorporate information gain also from mobile measurements to build autonomous systems capable of high throughput phenotyping. 

This paper makes the following contributions:
\begin{itemize}
    \item We present an active learning algorithm with a Gaussian Process model that incorporates the two types of measurements with different noise levels and determines high utility samples in order to actively learn the distribution of phenotypes in the field. (Section \ref{sec:gp})
    \item We propose an informative planning algorithm to determine the path with maximum joint information gain resulting from both static and mobile measurements. (Section \ref{sec:ipp})
    \item On a sorghum phenotype dataset collected by a robotic platform from a field in South Carolina, USA, we perform a detailed comparative analysis of our proposed algorithm with the current practices in crop phenotyping. (Section \ref{sec:data_collection})
    \item We have open-sourced our code repository, simulation environment and the sorghum dataset\footnote{Our github repository can be found at \url{https://github.com/sumitsk/algp.git}} for the research community to carry out further work in this direction.  
\end{itemize}

\section{Related Work}
Gaussian process~\cite{rasmussen2004gaussian} is a probabilistic non-parametric method widely used for modelling a scalar phenomenon as a multivariate normal distribution over the observable features. By utilizing the uncertainty estimate of this Bayesian model, many approaches have been proposed to address the problem of selecting a subset of available samples with the highest utilities to efficiently learn the scalar field~\cite{krause2008near, guestrin2005near}. Planning paths for a robot to navigate to those sampling locations is called Informative Planning~\cite{binney2013optimizing, ma2017informative, marchant2012bayesian, marchant2014sequential, lawrance2017fast}. Several metrics have been used to quantify the information gain of a sample or a set of samples, such as mutual information~\cite{luo2016distributed}, entropy~\cite{lu2014autonomous}, Fisher information~\cite{levine2010information} and the average reduction in variance~\cite{binney2012branch}. In this work, we use the entropy as the information-theoretic criteria.

Informative planning and adaptive sampling are widely studied topics in the robotics and machine learning communities. Informative planning algorithms based on a recursive-greedy style ~\cite{singh2007efficient, meliou2007nonmyopic} have been proposed with a sequential allocation mechanism in order to obtain subsequent waypoints. Low et al.~\cite{low2009multi} presented an entropy based framework which uses dynamic programming to obtain a batch of waypoints for determining the optimal path. Ma et al.~\cite{ma2016information} also proposed a similar informative planning method based on dynamic programming in order to compute the informative waypoints. All these approaches consider a single source of data acquisition and has the same measurement uncertainty for all the samples. Here, we consider two types of measurements with different noise levels resulting in a data-dependent noise model and seek to examine how the noisy data affects the ability of the agent to learn the environmental model. 
Note that our proposed method is directly applicable in situations where the robot has two sources of data acquisition - an expensive but accurate primary source and a noisier secondary source. For example, a robot equipped with the task of mapping the distribution of temperature in an indoor environment can take accurate thermometer readings from some locations and can also get temperature estimates from images captured by on-board cameras using deep learning techniques~\cite{chu2018visual}. Similarly, in crop phenotyping, penetrometers can be used for accurate measurement of stalk width and images can also provide the same estimates with the help of computer vision techniques~\cite{baweja2018stalknet}.

Binney at al.~\cite{binney2012branch} proposed a branch and bound algorithm for informative planning in a grid environment where each cell is a sampling location similar to the setup considered in this work. However, their proposed algorithm performs exhaustive search over a finite horizon and hence is limited to only small grids. In this work, we decompose the informative planning problem into two parts where the robot first selects some discrete locations to collect high quality measurements (\textit{static} samples) and then plans a path to maximize information gain from comparatively noisier measurements (\textit{mobile} samples) along the path too. As a result, our method can be applied to much larger environments than the ones considered in their work. 

Mueller et al.~\cite{mueller2017robotanist} presented a robotic platform for efficient crop phenotyping. The robot is capable of collecting physiological and morphological traits of crops. For example, the robot can measure the number and width of stalks, the height of plants, the area and color of leaves, etc. in each plot. The collected data can be used in various ways such as evaluating crops based on physical characteristics and can also be relayed to geneticists who use it to validate hypotheses and determine future plant crosses. Although the robotic platform enables high-throughput and robust phenotyping compared to human labor, its efficiency can be significantly increased by collecting high utility data in a short time which will enable the deployment of such systems to large fields where exhaustive coverage is not possible. Also, the amount of data needed to analyze crop genetics can be significantly reduced by learning the distribution of phenotypes via GP models so as to estimate unobserved data as shown in this work. 

\section{Methodology}
In this section, we describe our Gaussian Process model for incorporating both static and mobile measurements and informative planning algorithm to find the optimal path in the environment obtained by maximizing the joint information gain from both types of samples. 

\subsection{Gaussian Process Model} \label{sec:gp}
Let $V=\{v_{1},\dots,v_{n}\}$ be the set of all sampling data points in the environment where $v_{i} \in \mathbb{R}^{d}$ represents an observable feature vector. There exists a latent function $f:\mathbb{R}^{d} \rightarrow \mathbb{R}$ that maps the input $v \in \mathbb{R}^{d}$ to the objective value $f(v)$. In our work, the agricultural field is divided into plots and each plot represents a data point. The feature vector $v$ comprises of location, vegetation index and mean leaf angle density of the plants in a plot whereas the target function $f$ is the mean stalk height in the plot. Formally, $v = \{\text{location}, \text{vegetation index}, \text{leaf angle density} \}$ and $ y = \{\text{stalk height}\}$. In Section \ref{sec:exp}, we give a detailed reasoning behind selecting these features as input and output variables of the GP model. 

After sampling a set of data points $D \subset V$ and observing the corresponding target values $Y = \{y(v) \ | \ v \in D\}$, where $y(v)$ is a noisy estimate of the true value $f(v)$, the robot uses Gaussian Process regression to learn the underlying mapping $f$ assuming the joint distribution of the observed readings is Gaussian. A GP is a distribution over functions fully defined by a mean function $m$ and a covariance function $k$:  
\begin{align}
   f \sim \mathcal{GP}(m,k)  
\end{align}
The mean function $m$ is typically assumed to be zero without any loss of generality. The covariance function $k$, also known as kernel, encodes assumptions about the structure of the latent function $f$. It describes the relation between two data points and typically has some free hyperparameters to control this relation. We used the popular Matern~\cite{minasny2005matern} Kernel as our covariance function:
\begin{align} \label{eq:matern}
    k_{\text{Matern}}(x,x') = \sigma^2\frac{2^{1-\nu}}{\Gamma(\nu)}\left(\frac{\sqrt{2\nu}|x-x'|}{l}\right)^\nu K_\nu\left(\frac{\sqrt{2\nu}|x-x'|}{l}\right)
\end{align}
where $\Gamma$ is the gamma function, $K_{\nu }$ is the modified Bessel function, and $\sigma$ and $l$ are the output-scale and the length-scale parameters of the kernel respectively. In this work, we used $\nu=1.5$. 

Since there are two types of samples here, the associated measurement noise is not same for all the data points but instead depends on the type of sampling - \textit{static} or \textit{mobile}. We denote the variances associated with static and mobile measurements as $\sigma_{s}^{2}$ and $\sigma_{m}^{2}$ respectively. One can see that $\sigma_{m}^{2}$ depends on the quality of images captured while the robot is in motion which in turn depends on the speed of robot. For simplicity, we assume that the robot moves at a uniform speed and hence $\sigma_{m}^{2}$ is constant. Furthermore, $\sigma_{s}^{2}$ is also a constant as the robot is at rest while collecting static samples in order to minimize measurement errors and get high quality data. 

As the robot collects mobile data while moving in the environment, it is possible that it has acquired multiple measurements for the same feature vector $v$. This happens when the robot travels through a region it has visited before. Since Gaussian Processes are non-parametric method, there can not be two different target values for the same input $v$. We combine all the mobile measurements into a single equivalent measurement by taking their mean value:
\begin{align*}
    \hat{y}_{m}(v) &= \frac{1}{n_{m}(v)}\sum_{k=1}^{n_{m}(v)} y_{m}^{k}(v)
\end{align*}
where $n_{m}(v)$ is the number of mobile measurements for data point $v$ and $\{y_{m}^{k}(v)\}_{k=1}^{n_{m}(v)}$ are the observed values. This averaging step reduces the noise in the observed data by biasing it towards the true mean value. Furthermore, a robot may have acquired a static measurement $y_{s}(v)$ for a same data point $v$ for which it has gathered mobile measurement(s) before. Instead of substituting all the comparatively noisy mobile data with a single static data, we fuse them together as the product of the two probability density functions: 
\begin{align*}
    y(v) &= \frac{\frac{y_{s}(v)}{\sigma_{s}^{2}} + \frac{\hat{y}_{m}(v)}{\sigma_{m}^{2}}}{\frac{1}{\sigma_{s}^{2}} + \frac{1}{\sigma_{m}^{2}}}\\
\sigma^{2}(v) &= \frac{\sigma^{2}_{s}\sigma^{2}_{m}}{\sigma^{2}_{s} + \sigma^{2}_{m}}
\end{align*}
Essentially, the cumulative output value $y(v)$ is a weighted mean of the measurement from the two sources where the weights are the precision of each source and the resulting variance $\sigma^{2}(v)$ drops below the measurement variance of each of the sources. If a data point $v$ has been sampled by only one medium (\textit{static} or \textit{mobile}), then the associated data-dependent noise is either $\sigma_{s}^{2}$ or $\sigma_{m}^{2}$ depending on the type of sampling. The covariance due to the sample-dependent measurement noise can be represented by using a white noise kernel:
\begin{align}
  k_{w}(v_{i},v_{j}) = \sigma^{2}(v_{i})\delta_{ij} \label{eq:white_noise}
\end{align}
where $\delta_{ij}$ is the Kronecker delta function which is $1$ if $i=j$ and $0$ otherwise. We model the inter-sample covariance function as the sum of Matern covariance function and the white noise covariance function:
\begin{align} \label{eq:cov}
  \Tilde{k}(v_{i},v_{j}) &= k_{\text{Matern}}(v_{i},v_{j}) + k_{w}(v_{i},v_{j})
\end{align}
The hyperparameters of the model can be estimated by maximizing the log marginal likelihood of the observed data as done in ~\cite{rasmussen2004gaussian}. With the GP model fully defined, we can now write the posterior target distribution of a set of samples $A \subset V$ conditioned on the sampled set $D$ as:
\begin{align}
    f(A)\ |\ A,D,Y &\sim \mathcal{N}(\mu_{A|D}, \Sigma_{A|D})   \nonumber   \\
    \mu_{A|D} &= \Sigma_{AD}\Sigma_{DD}^{-1}Y   \nonumber \\
    \Sigma_{A|D} &= \Sigma_{AA} - \Sigma_{AD}\Sigma_{DD}^{-1}\Sigma_{DA}    \label{eq:ent}
\end{align}
where $\Sigma_{AB}$ is the pairwise covariance matrix whose $(i,j)^{th}$ element is the covariance between the $i^{th}$ sample of $A$ and the $j^{th}$ sample of $B$ as defined in Equation \ref{eq:cov}. 

The notion of differential entropy is often used in spatial statistic optimization problems to specify the informativeness of a set of samples given a set of previously collected samples. Formally, the entropy of a set $A$ conditioned on a sampled set $D$ is defined as:
\begin{align*}
    H\left(A|D\right) = \frac{1}{2}\log{\left(\left(2\pi e\right)^{|A|} \det\left(\Sigma_{A|D}\right)\right)}
\end{align*}
where the conditional covariance matrix $\Sigma_{A|D}$ is estimated from the posterior distributed as described in Equation \ref{eq:ent}.

\begin{figure*}[ht]
    \centering
        \begin{minipage}{0.35\textwidth}
        \centering
        \includegraphics[width=0.95\textwidth]{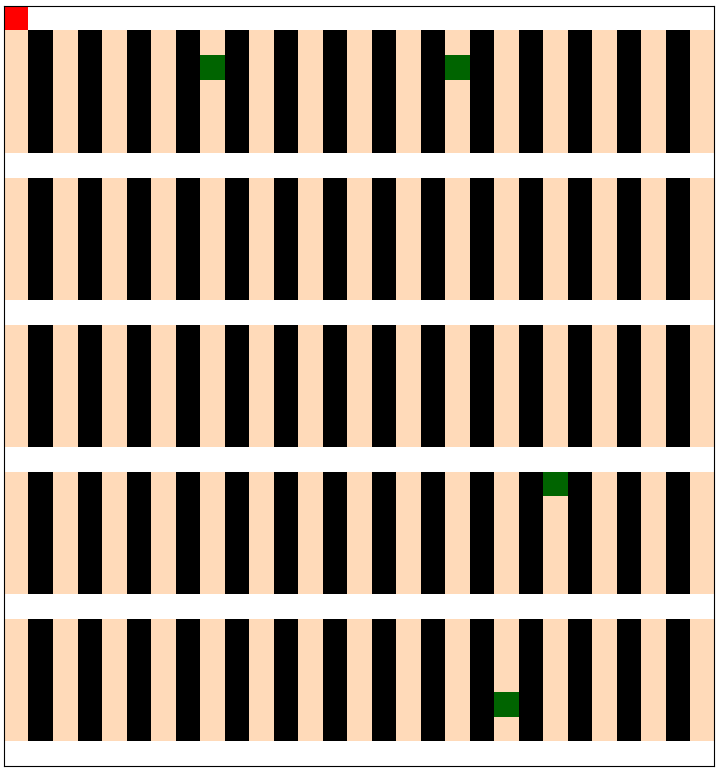} 
        \subcaption{Environment}
        \label{fig:static_sampling}
    \end{minipage}\hfill
    \begin{minipage}{0.65\textwidth}
        \centering
        \includegraphics[width=0.95\textwidth]{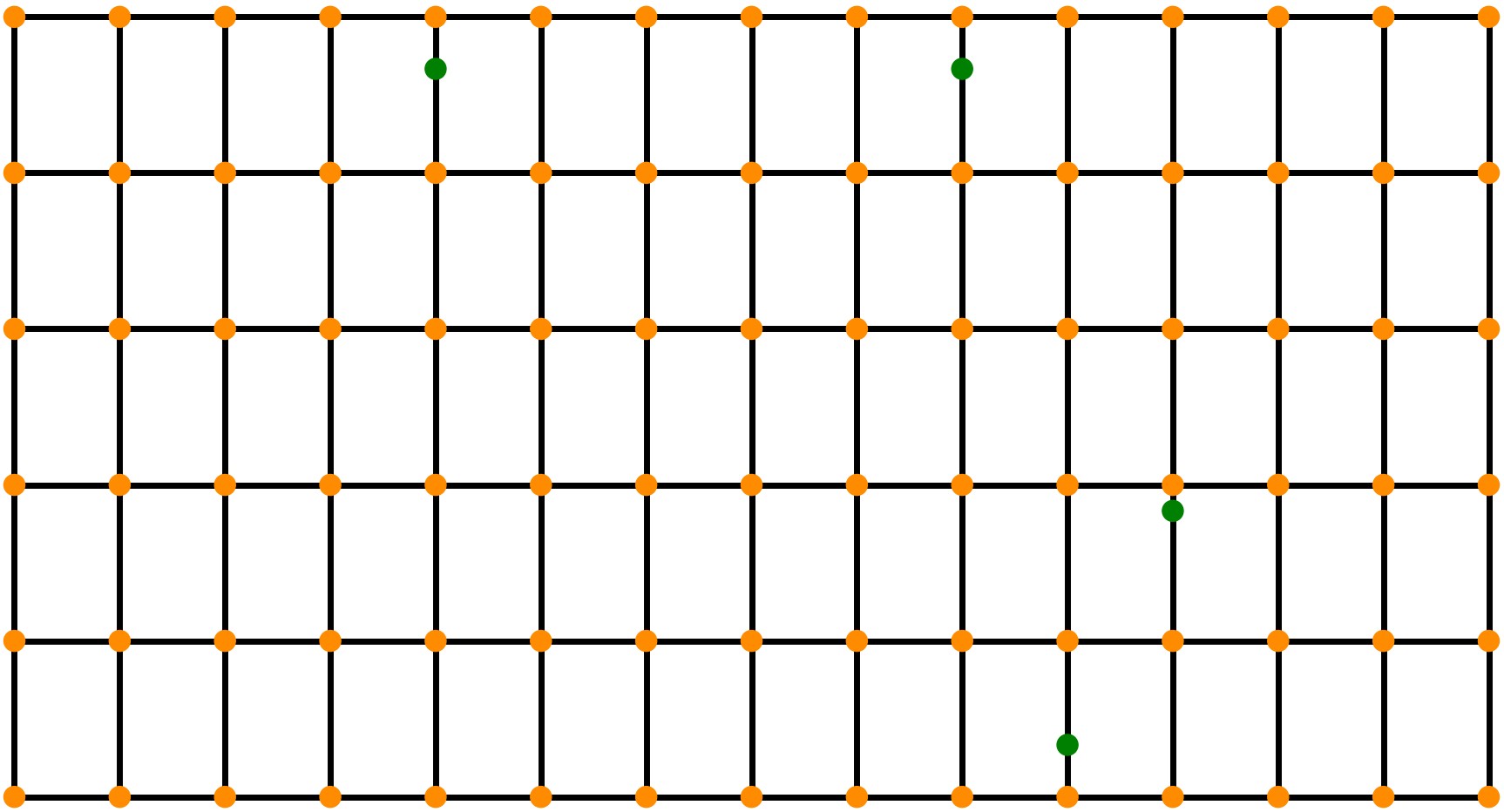} 
        \subcaption{Graph}
        \label{fig:graph_contruction}
    \end{minipage}
    \caption{(a) In the first step of Informative Planning, the robot determines the $p$ most informative samples (here $p=4$) in the field which are shown in green. (b) In the second step, it constructs a graph whose nodes are the entry and exit of all the narrow corridors in the field shown in orange plus the static sampling locations determined in the first step shown in green. Path planning in the original grid environment is equivalent to that on this graph $G$ however significantly faster. }\label{fig:graph}
\end{figure*}

\subsection{Informative Planning} \label{sec:ipp}
We now describe our proposed two-step informative planning algorithm which takes into account both static and mobile measurements to find the most informative path in the environment. In this work, we use differential entropy of a set of samples as its information gain metric.

In the first step, the robot selects a set $A^{*}$ of $p$ points (see Figure \ref{fig:iter1}) from the unsampled space with the maximum entropy conditioned on the already sampled set $D$. Formally, 
\begin{align}\label{eq:sensing_set}
A^{*} = \argmax_{A \in \mathcal{P}(V\setminus D), \ |A|=p} H(A|D)    
\end{align}
where $\mathcal{P}(S)$ is the power set of $S$. Since, this is an NP-Hard problem \cite{guestrin2005near}, we use the greedy strategy proposed by Krause et al.~\cite{krause2008near} where the $i^{th}$ sample $a_{i}$ is selected from the unsampled set as the one which results in the maximum information gain:
\begin{align}
a_{i}^{*} = \argmax_{a_{i} \in V \setminus (A_{i-1}^{*},D)} H\left(a_{i}|A_{i-1}^{*},D\right)    \label{eq:greedy_max}
\end{align}
where $A_{i-1}^{*} = \{a_{1}^{*},\dots,a_{i-1}^{*}\}$. The white noise associated with these samples is the static measurement noise, i.e., $\sigma^{2}(v) = \sigma_{s}^{2} \ \forall \ v \in A^{*}$.  

In the next step, the robot plans a path from its current location $x_{0} \in \mathbb{R}^{2}$ (see Figure \ref{fig:iter1}) to collect data from all the selected data points $A^{*}$ while maximizing the joint information gain from both static and mobile samples encountered in the path. 

Let, $\Omega_{A,x}$ be the set of all possible paths originating from a location $x \in \mathbb{R}^{2}$ in the environment and passing through the plot of all samples $a \in A$. Also, let $P_{m}$ be the set of mobile samples along the path $P$. The robot determines the most informative path $P^{*}$ while satisfying the budget constraints: 
\begin{align*}
    P^{*} = \argmax_{P \in \Omega_{A^{*},x_{0}} , \ c(P)<B} H\left(A^{*},P_{m}|D\right)
\end{align*}
where $c(P)$ is the cost of the path considered to be the length here and $B$ is the budget or the maximum allowed path length. Using the chain rule of entropy, we can write: 
\begin{align}
    P^{*}= \argmax_{P \in \Omega_{A^{*},x_{0}} , \ c(P)<B} H\left(P_{m} | A^{*},D\right) \label{eq:optimization}
\end{align}

The robot aims to find the path which results in the maximum information gain from mobile samples $P_{m}$ conditioned on the previously sampled set $D$ and the static samples $A^{*}$. This is expected as all the candidate paths have the same set of static samples so the best path is the one which results in maximum conditional information gain from mobile samples. Normally, the path optimization criteria in adaptive sampling applications is the cost of travel and once a set of sampling locations is determined, the planner generates the shortest route to traverse all of them~\cite{ma2017informative}. However, in our case, due to the presence of an additional medium of gathering data, the robot seeks to find the path which results in \textit{joint maximum information gain} from both the mediums. 

We set the budget $B$ to be the shortest path length $c_{\text{min}}$ plus a slack term $\xi$:
\begin{align}\label{eq:budget}
c_{\text{min}} &= \min_{P \in \Omega_{A^{*},x_{0}}} c(P) \nonumber \\
 B &= c_{\text{min}} + \xi  
\end{align}

This adaptive budget ensures that there exists at least one path which satisfies all the required conditions. The slack term $\xi$ controls the freedom given to the robot to explore areas not necessarily along its shortest path. If we enforce a constant budget instead, the robot may go to far off places if the shortest path length is much smaller than the budget. Also, there may not be any feasible path if the budget is inadequate. Moreover, in the adaptive sampling cases where the robot is required to do multiple traversals, having a budget that ensures the existence of at least one feasible path while also providing some freedom to explore distant areas is desirable. 

Since there is very narrow space between two rows of crops in the field, it is not possible for the robot to take a \ang{180} turn without damaging the plants. This means that once the robot enters the narrow region (corridor) between two rows of crops, it can only move in its heading direction till it reaches the end of it where it can either take a \ang{90} turn or keep moving forward (see Figure \ref{fig:static_sampling}). This no U-turn constraint significantly simplifies the planning problem by reducing the search space from the entire field to a small graph $G$ as shown in Figure \ref{fig:graph_contruction}. The nodes of this graph are the intersection points or junctions in the environment and edges are the straight line paths connecting them. To find the optimal path from the robot's initial location $x_{0}$ traversing through the set of location of all the static samples $a_{i}^{*} \in A^{*}$ denoted by $W = \{w_{i} = \text{loc}(a_{i}^{*}) | \ a_{i}^{*} \in A^{*} \}$, we add the initial position of robot $x_{0}$ and the waypoints $W$ to the graph $G$. One can easily see that path planning on this graph is equivalent to that on the original grid environment but highly efficient.

The robot runs a graph search on $G$ to determine the set of potential paths $\Omega_{A^{*},x_{0}}$ which satisfy the budget constraints and then selects the optimal path $P^{*}$ as per equation \ref{eq:optimization}. As is common in path planning, we use a heuristic to approximate the goal distances to focus the search in order to obtain solutions faster than uninformed search method~\cite{nilsson1971problem}. In our case, there is no explicit goal node but a goal state achieved when all the waypoints have been visited. 

We use a simple method to find a lower bound on the distance to cover all the remaining sampling locations. Let, the current position of the robot and the waypoints left to be visited be $x = \{x^{1},x^{2}\} \in \mathbb{R}^{2}$ and $W$ respectively. We denote $W^{1}=\{w^{1}|w\in W \}$ and $W^{2}=\{w^{2}|w\in W \}$. We compute a lower bound on the path cost by determining the bounding box formed by $x$ and the set of waypoints $W$. Note that the agent's path will touch or intersect all $4$ edges of this bounding box in order to visit all the remaining waypoints. In other words, the agent has to travel at least the distance from its current position to the nearest edge and then to the opposite edge along each of the two axes. Formally, the proposed heuristic cost-to-go $h(x,W)$ can be written as:
\begin{align*}
    h(x,W) = \ &\text{max}\left(\{x^{1},W^{1}\}\right) - \text{min}\left(\{x^{1},W^{1}\}\right) \\ 
    + & \text{max}\left(\{x^{2},W^{2}\}\right) - \text{min}\left(\{x^{1},W^{1}\}\right) \\
    + & \text{min}\left(\text{max}\left(\{x^{1},W^{1}\}\right) - x^{1}, x^{1} - \text{min}\left(\{x^{1},W^{1}\}\right)\right) \\
    + & \text{min}\left(\text{max}\left(\{x^{2},W^{2}\}\right) - x^{2}, x^{2} - \text{min}\left(\{x^{2},W^{2}\}\right)\right)
\end{align*}
The first two terms are the dimensions of the bounding box whereas the last two terms are the distance of the agent from the nearest edge along each axes. This lower bound on the cost-to-go can be further tightened by incorporating the no \ang{180} turn constraint and the initial orientation of the agent in the analysis, however, for the size of environment considered in this work, the proposed heuristic was able to significantly speed up the search and provide optimal plans in real-time. 

In this work, we exhaustively compute all the paths which satisfy the budget constraints, however, this can be computationally very expensive as the size of the environment increases. For the crop field considered here, exhaustive evaluation was computationally tractable. For larger environments, however, this step can be the computational bottleneck in the entire pipeline. One way to solve this problem is to construct the desired path in segments. Instead of determining all possible paths from the start location at once, one can first find an order of visiting waypoints by minimizing the total travel cost, for example, and then select the optimal path between two consecutive waypoints in the determined order.

\section{Data collection} \label{sec:data_collection}
\begin{figure}
    \begin{minipage}{.45\textwidth}
        \centering
        \includegraphics[width=.8\textwidth]{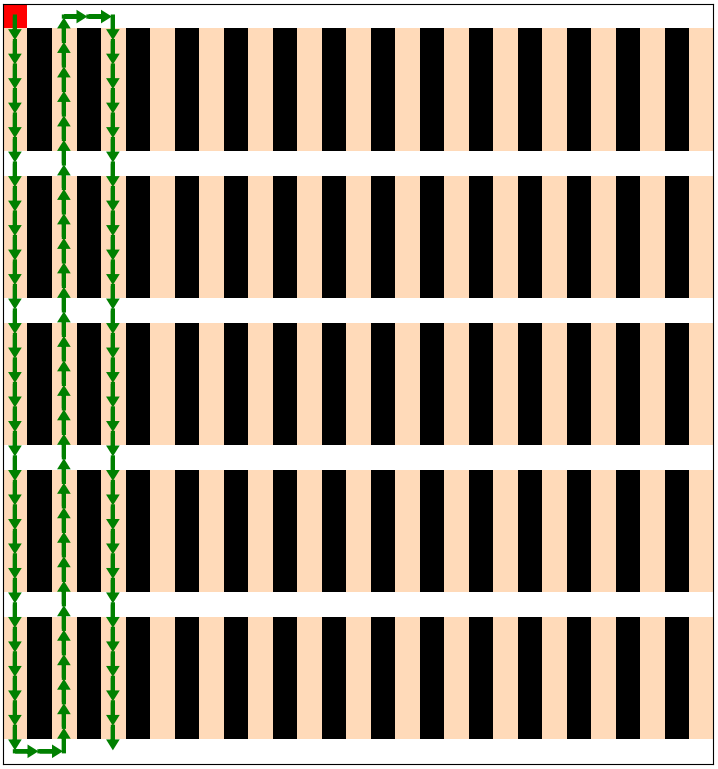}
        \caption{The figure shows the data collection process. The robot (shown in red) takes phenotypes measurements from all the plots in the first column. After that, it shifts to the next free column containing an array of plants and continues in the same manner till it has covered the whole field.}
        \label{fig:naive_strategy}
    \end{minipage}\hfill
    \begin{minipage}{.5\textwidth}
        \centering
        \includegraphics[width=\textwidth]{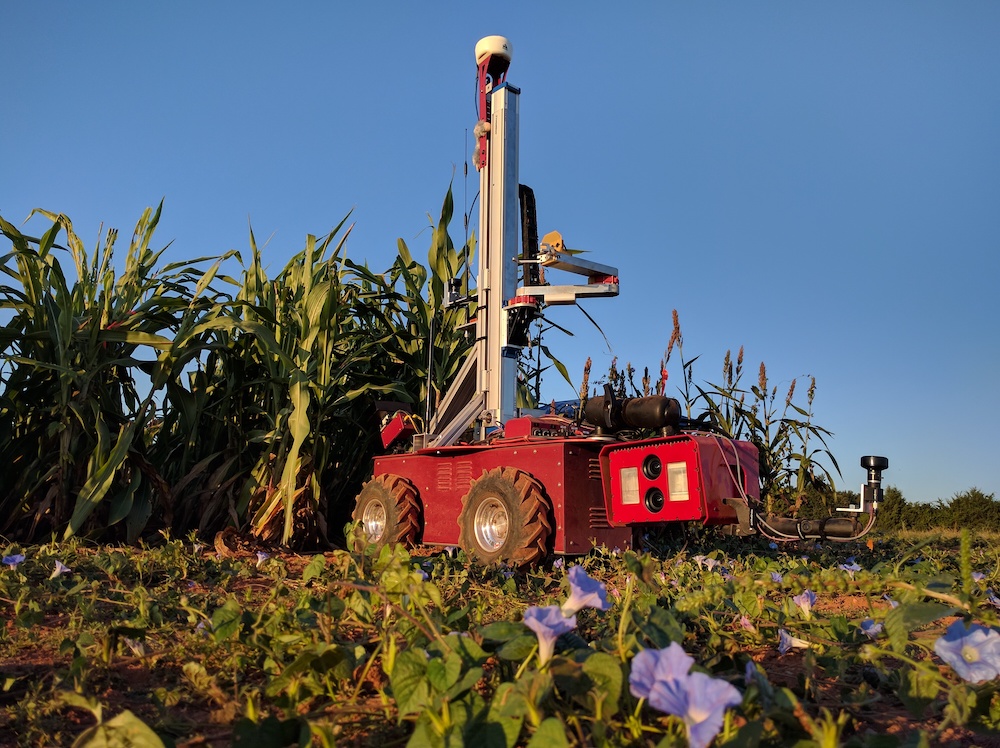} 
        \caption{Our robotic platform Robotanist in a sorghum field in South Carolina, USA.}
        \label{fig:robot}
    \end{minipage}
\end{figure}

The phenotype measurements are collected by a robotic platform (see Figure ~\ref{fig:robot}) from a sorghum field in South Carolina, USA. The sorghum field we are studying is laid out in the form of a grid of $25\times15$ plots. For taking phenotype measurements, the robot exhaustively visits all the plots from North to South (see Figure \ref{fig:naive_strategy}) and captures images in each plot from its on-board cameras. After reaching the end of a column, it shifts two columns East and then moves from South to North till it reaches the end. The robot continues this process till it has covered the whole field. As one can see, this exhaustive coverage strategy is time and fuel consuming as the robot has to stop or move slowly in each plot to get accurate measurements. The objective of this work is to develop an intelligent and adaptive strategy to enable fast data acquisition for further use by scientists to validate genetic hypotheses and determine desirable plant crosses. 

\begin{figure}[ht]
    \begin{minipage}{.25\textwidth}
        \centering
        \includegraphics[width=\textwidth]{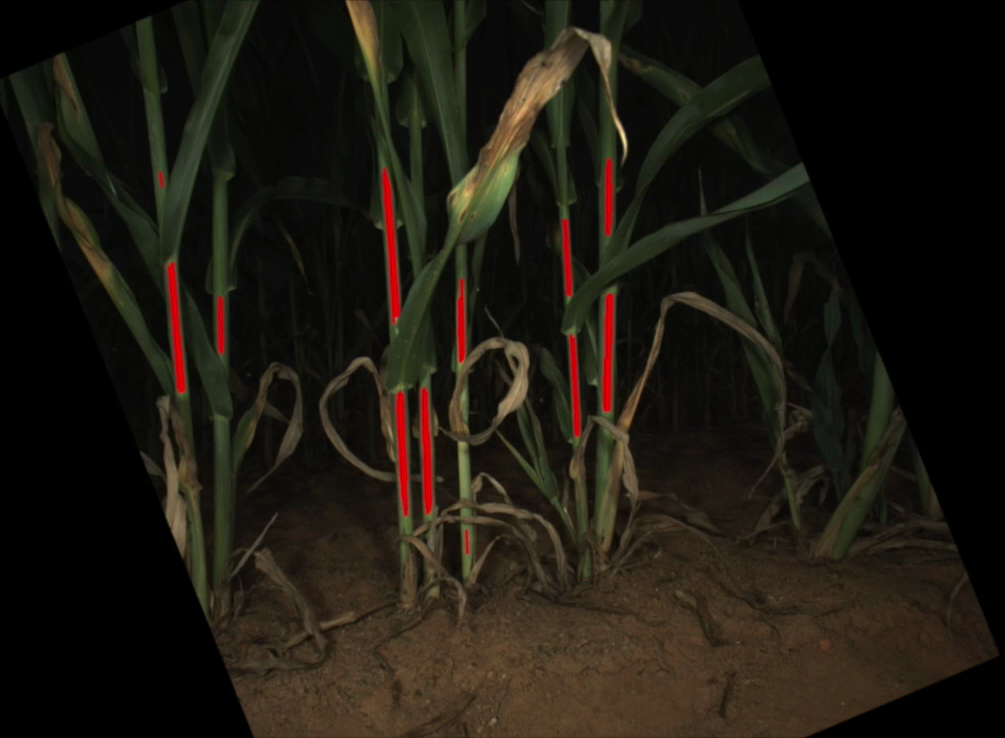} 
        \subcaption{Detected stalks}
        \label{fig:stalk}
    \end{minipage}\hfill
    \begin{minipage}{.21\textwidth}
        \centering
        \includegraphics[width=\textwidth]{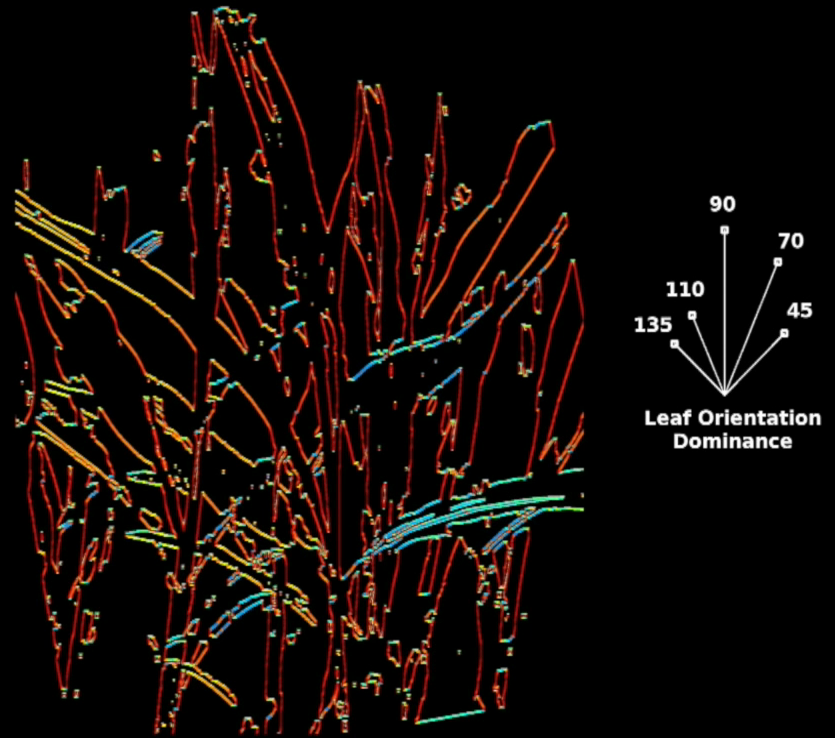} 
        \subcaption{Leaf angle}
        \label{fig:leaf_angle}
    \end{minipage}\hfill
    \begin{minipage}{.26\textwidth}
        \centering
        \includegraphics[width=\textwidth]{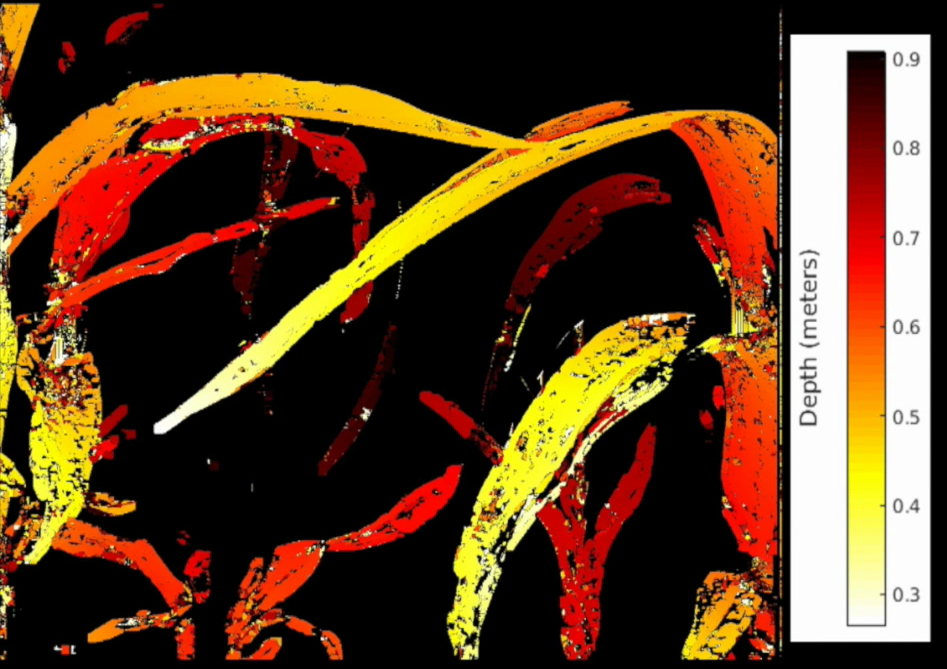} 
        \subcaption{Leaf area}
        \label{fig:leaf_area}
    \end{minipage}\hfill
    \begin{minipage}{.23\textwidth}
        \centering
        \includegraphics[width=\textwidth]{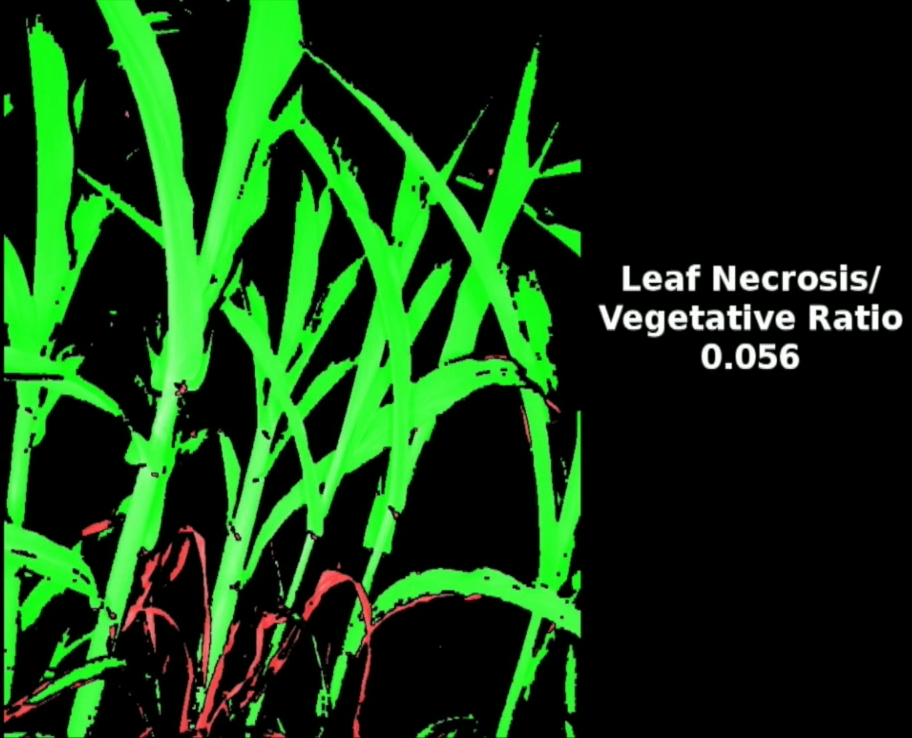} 
        \subcaption{Vegetation index}
        \label{fig:vi}
    \end{minipage}
    \caption{ (a) The robot detects stalks and measures their width, height and count in each plot from images. Similarly, (b) leaf angle, (c) leaf area and (d) vegetation index or the ratio of dry to green leaf area are measured from the captured images.}\label{fig:data_collection}
\end{figure}

As the robot drives through the field, it measures the width, height and the number of stalks in each plot from the images of stalk regions of plants as shown in Figure \ref{fig:stalk}. It also measures the leaf angle, surface area of leafs and the vegetation index which is the ratio of dry to green leaf area from the images of leafs as shown in Figures \ref{fig:leaf_angle}, \ref{fig:leaf_area} and \ref{fig:vi} respectively. We have developed sophisticated deep learning and computer vision techniques in our previous works to extract these measurements directly from images. 

We are interested in studying these phenotype measurements and determining their correlation with the yield produced at the end of the harvest season. With the help of such a correlation model, scientists will be able to determine early in the season which genetic varieties of a crop are expected to give high yield and need not have to wait till harvest to obtain the results of their experiments. Furthermore, by actively taking high utility measurements from some plots only and not the entire field, the data acquisition process can be accelerated and the research work can be further extended to large agricultural fields. 


\section{Analysis}
In this section, we describe our simulation environment, implementation details of our algorithm and the experiments we performed to evaluate the efficacy of our proposed algorithm. 

\subsection{Implementation}
For rapid testing and experimentation purposes, we have built a grid-based simulation environment (shown in Figure \ref{fig:env}) which imitates the actual layout of plots in the sorghum field. Each obstacle-free column represents an array of crops planted with intermittent spacing allowing the robot to shift to an adjacent array. The measurement received by a robot in simulation is drawn from a normal distribution with mean as the ground truth value and variance as either static variance or mobile variance depending on the type of sampling. Formally, 
\begin{align*}
    y_{s}(v) &\sim \mathcal{N}\left(y(v), \sigma_{s}^{2}\right)\\
    y_{m}(v) &\sim \mathcal{N}\left(y(v), \sigma_{m}^{2}\right)
\end{align*}
where $y_{s}(v)$ and $y_{m}(v)$ are the static and mobile measurements respectively and $y(v)$ is the ground truth field data for the sample $v$. 

Our GP model as described in Section \ref{sec:gp} is built on top of GPyTorch~\cite{gardner2018gpytorch} which is a Gaussian Process library implemented using PyTorch~\cite{paszke2017automatic}. We used zero mean and Matern covariance function with $\nu=1.5$ (see Equation \ref{eq:matern}) as the mean and covariance modules of the GP model respectively. The hyperparameters of the GP model are learned by maximizing the log likelihood of a subset of training data and are fixed thereafter. The informative planning and graph search algorithms are implemented with the help of NetworkX ~\cite{hagberg2008exploring} package. Our simulation environment, code repository and the sorghum dataset are open-sourced and can be found at \url{https://github.com/sumitsk/algp.git}.

\subsection{Experiments}    \label{sec:exp}
Since true yield data will be available only after the harvest season, we used the mean stalk height in each plot as a metric for quality of produce from that plot. The density of crops is more or less same throughout the field, so the average plant height in a plot directly reflects the expected yield at harvest. The agent aims to actively learn the distribution of mean stalk height as a function of location, mean vegetation index and mean leaf angle density using a GP model. We included the last two features in the set of input variables as sorghum phenotypes are poorly correlated with location as suggested by crop scientists and no meaningful mapping can be learned between them. Both leaf angle and vegetation index are measured from the images of leafs which can be captured by aerial vehicles from the top in a short time. So, we reasonably assumed that the ground robot has access to mean vegetation index and mean leaf angle density in each plot prior to starting its adaptive sampling routine. 

In each simulation, the robot starts from the top left corner of the grid as shown in Figure \ref{fig:grid_env} and performs a total of $8$ iterations. In each iteration, it determines $p=4$ most informative static sampling locations and the corresponding optimal path (see Figure \ref{fig:iter1}). After collecting all the data from the planned path, it updates its GP model. The robot repeats this process in the next iteration (see Figure \ref{fig:iter2}). 

\subsubsection{\textbf{Comparison with baselines}}
We compared our proposed Maximum Entropy algorithm (\textit{MaxEnt}) as described in Section \ref{sec:ipp} against four baseline methods:
\begin{itemize}
\item \textit{Naive static}: The robot sequentially visits all the plots in each column as shown in Figure \ref{fig:naive_strategy}. In order to get accurate measurements, the robot slows down or stops in each plot. In this case, the white noise variance is equal to the static measurement variance for all the samples, i.e.,  $\sigma^{2}(v) = \sigma_{s}^{2}$ $\ \forall v\in V$ .
\item \textit{Naive mobile}: This strategy is same as \textit{Naive static} except that the agent does not stop in each plot and measures phenotypes while moving. The data collected in this process is noisier than that in the previous strategy. Here, the white noise variance for all the samples is the mobile measurement variance, i.e., $\sigma^{2}(v) = \sigma_{m}^{2}$ where $\sigma_{m}^{2}> \sigma_{s}^{2}$.
\item \textit{Shortest}: Like \textit{MaxEnt}, in this case, the agent determines the set of all the feasible paths $\Omega_{A^{*},x_{0}}$ as described in Section \ref{sec:ipp}. It then selects the path with the shortest length. 
\item \textit{Equi-sample}: After computing the set of all the feasible paths $\Omega_{A^{*},x_{0}}$, the agent selects the path which has the same number of sampling locations as the one selected by \textit{MaxEnt} algorithm.  
\end{itemize}

Note that \textit{MaxEnt}, \textit{Shortest} and \textit{Equi-sample} are informative strategies, i.e., they adaptively select the next sensing locations and plan paths to maximize information gain instead of naively covering the entire field as done by \textit{Naive static} and \textit{Naive mobile} strategies. In \textit{Shortest} and \textit{Equi-sample} strategies, ties among multiple candidate paths are broken arbitrarily. 

We ran $20$ simulations to compare the $5$ strategies. In each simulation, out of $375$ sensing locations, we reserved a set of randomly chosen $40$ locations as a test set denoted by $V_{T}$. The robot can not observe this test set while learning the distribution of target variable. The metric used to quantify the performance of a strategy is Mean Absolute Error (MAE) between the true phenotype values and the ones predicted by the model on the test set samples: 
\begin{align*}
    MAE = \frac{1}{|V_{T}|}\sum_{v \in V_{T}} \left| \mu(v) - y(v)\right|
\end{align*}
where $\mu(v)$ is the predicted mean value by the GP model and $y(v)$ is the ground truth phenotype value collected from the field. 

For this experiment, we set $\sigma_{s}=0.5$, $\sigma_{m}=2.5$ and $\xi=0$. We would like to mention that the true values of stalk width varies from $20$ cm to $85$ cm and we did not perform any normalization on the dataset.

\begin{figure}
    \begin{minipage}{.5\textwidth}
        \centering
        \includegraphics[width=\textwidth]{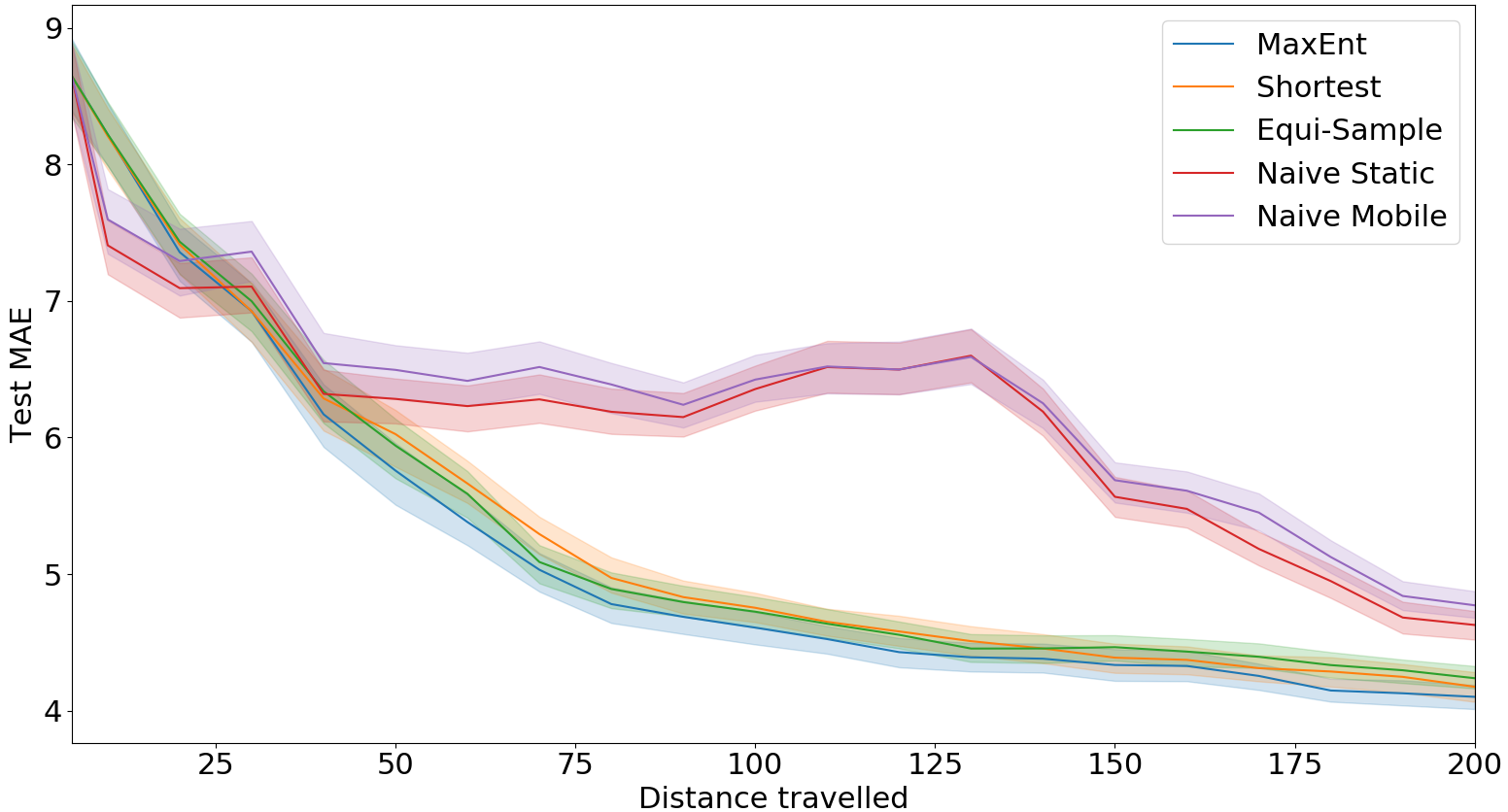} 
        \subcaption{Test MAE v/s distance}
        \label{fig:mae_vs_dist}
    \end{minipage}\hfill
    \begin{minipage}{.5\textwidth}
        \centering
        \includegraphics[width=\textwidth]{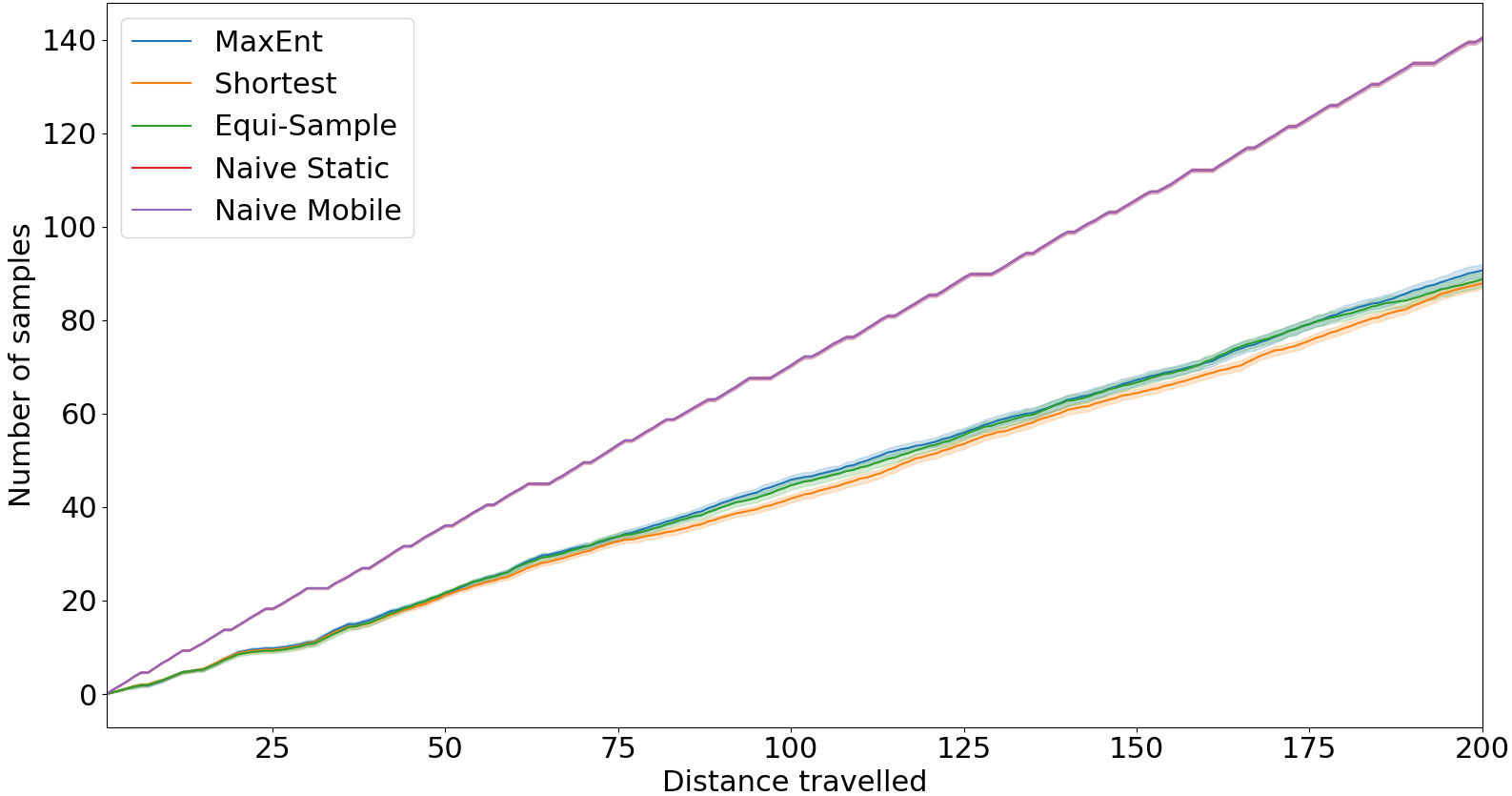} 
        \subcaption{Number of samples v/s distance}
        \label{fig:count_vs_dist}
    \end{minipage}
    \caption{The plot shows the (a) MAE of agent's prediction on the test set against distance travelled and (b) number of samples collected against distance travelled. For clarity of the picture, only $50\%$ confidence interval is shown.}
    \label{fig:strategies}
\end{figure}

The graphic plot of agent's prediction against distance travelled is shown in Figure \ref{fig:mae_vs_dist}. Each grid cell in the environment corresponds to 1 unit distance. Note that for the same distance, the informative strategies gather less number of samples than the naive ones (see Figure \ref{fig:count_vs_dist}) as the planned path often requires travelling along a row in the grid (see Figure \ref{fig:iter1}) where there are no sampling sites. On the other hand, the naive strategies move to the next column only after sampling all the plots in the current column (see Figure \ref{fig:naive_strategy}). In each iteration, the agent collects a variable number of samples, so we decided to use the distance covered as our independent axis in the plot in order to ensure fair comparison among all the strategies.

Unlike the naive strategies, all $3$ informative strategies are able to quickly predict the phenotype distribution by actively visiting places with high utility. Our proposed \textit{MaxEnt} consistently achieves the lowest prediction error on the test set indicating its ability to accurately estimate the target distribution. \textit{Shortest} and \textit{Equi-sample} strategies are also able to learn the target distribution and closely match the performance of \textit{MaxEnt}. On the other hand, the two naive strategies perform poorly and even after covering more than half of the field and collecting more samples that the informative strategies, they are unable to match the predictive performance of the informative strategies. The experiment validates that the agent can learn the target distribution in a short time by actively sensing locations and does not need to exhaustively cover the entire field and collect data from all the plots as done in the current practices. 

\subsubsection{\textbf{Effect of noise ratio}}
We define the noise ratio $k$ as the ratio of mobile to static measurement noise, i.e., $k = \sigma_{m} / \sigma_{s}$. Assuming a fixed static measurement noise $\sigma_{s}$, the noise ratio depends on the speed of the robot. The faster the motion of robot,  the larger is the noise ratio and vice-versa. This is because if the robot is moving fast, then the measurement uncertainty will increase due to increase in localization errors and poor image quality. However, moving too slow is also undesirable as it slows down the data collection process, hence there is a trade-off between the quality of collected data and the time consumed in the process. 

We compare the predictive performance of \textit{MaxEnt} on the test set samples against different values of noise ratio. For this experiment, we set $\sigma_{s}=0.5$ and $\xi=0$. We ran $20$ simulations and the results obtained are shown in Table \ref{tb:noise_ratio}.
\begin{table}[h!]  
\centering
\begin{tabular}{c c c c c}
\hline
\multicolumn{5}{c}{\textit{MaxEnt} prediction with different $k$}\\
& $k=1$ & $k=2$ & $k=5$ & $k=10$\\
\hline
MAE & 4.08(0.70) & 4.10(0.77) & 4.09(0.78) & 4.40(0.68)\\
\hline
\end{tabular}
\caption{MAE on the test set for $\sigma_{s}=0.5$ and $\sigma_{m}=k\sigma_{s}$ averaged over $20$ simulations. In each simulation, the robot travels $250$ distance units. The first term is the mean whereas the one inside the parentheses is the standard deviation of the metric.}\label{tb:noise_ratio}
\end{table}
\vspace{-20pt}

We observe that there is not much significant difference in the predictive accuracy of the learned model till $k=5$. This is an interesting finding as it suggests that the robot does not need to stop or move extremely slowly in each plot in order to learn an accurate model of the distribution of phenotypes in the field but an intelligent and adaptive combination of static and mobile samples can also achieve the same level of accuracy. Since, noise ratio depends on speed, one can estimate the speed at which the robot should move in the field in order to collect samples rapidly without deteriorating the quality of learned GP model. However, there are many other factors that affect the measurement noise like terrain complexity, camera quality, weather conditions, etc. and an in-depth analysis of their effect along with the speed of robot is left for future work. 

\subsubsection{\textbf{Effect of budget}}
In the previous experiments, we set $\xi = 0$ which means that the robot does not have any extra budget to explore nearby areas in each iteration. We would like to mention that even in this case, there exists many candidate paths of the same length because of the structured warehouse-like layout of the environment. Here, we want to understand how does some slack on the maximum allowed path length or extra budget affects the quality of the learned model. Increasing slack does provide some freedom to the robot to explore areas not necessarily along its shortest path, however, that comes at the expense of increased computational cost. Hence, it is necessary to determine how much additional budget, if any, is useful in adaptive crop phenotyping applications while maintaining tractable computational requirements. 

We compare the final predictions made by \textit{MaxEnt} algorithm on the samples in the test set for different values of slack length $\xi$. We set $\sigma_{s}=0.5$ and $\sigma_{m}=2.5$ for this experiment and ran $20$ simulations. The results are shown in Table \ref{tb:slack}. 

\begin{table}[h!]  
\centering
\begin{tabular}{c c c c c}
\hline
\multicolumn{5}{c}{\textit{MaxEnt} prediction with different slack $\xi$}\\
& $\xi=0$ & $\xi=5$ & $\xi=10$ & $\xi=15$\\
\hline
MAE & 4.22(0.58) & 4.00(0.51) & 4.08(0.54) & 4.19(0.40)\\
\hline
\end{tabular}
\caption{MAE on the test set for $\sigma_{s}=0.5$ and $\sigma_{m}=2.5$ for different slack values averaged over $20$ simulations. In each simulation, the robot travels $250$ distance units. The first term is the mean whereas the one inside the parentheses is the standard deviation of the reported metric.}\label{tb:slack}
\end{table}

We observe an interesting trend in the final predictions made by the learned model. Increasing the slack value improves the performance of the model, however, the increase is not monotonic, i.e., providing additional budget does not necessarily results in increased performance of the learned GP model for the same distance travelled. This is because samples lying nearby along a path exhibit high correlation due to the local smoothness modelling assumptions of the GP kernel, hence, the information gained along a path does not always increase with increase in path length.  As a result, a large budget does not necessarily lead to large improvement in model performance although it significantly increases the total computational cost of evaluating feasible paths. On the other hand, a small extra budget is sufficient for the agent to gather any useful additional information helpful in learning an accurate model of the target distribution while maintaining a reasonable computational load.

\section{Conclusion}
We presented an active learning framework that alternates between adaptively sampling plots with high utility and learning a GP model of the target distribution for High Throughput Phenotyping. We have also released our code repository for further work by other researchers. Through simulation experiments, we demonstrated the superior performance of our proposed approach compared to the current practices. We also performed several ablation experiments to understand the contribution of various components in the whole planning and learning pipeline. 

We believe this is just the beginning and there is a lot of work to be done for developing reliable robotics and machine learning technologies to accelerate the breeding process. Here, we used only one phenotype (mean stalk height) as an indicator of crop health. In future work, we will analyze correlations between multiple phenotypes by modelling them in a Multi-task Gaussian Process framework for active learning and efficient data gathering. Also, we will deploy a team of robots working in a fully decentralized manner to adaptively estimate the quality of produce. 

\bibliographystyle{unsrt}
\bibliography{references}
\end{document}